\documentclass{article}
\pdfoutput=1

\usepackage[preprint]{neurips_2024}

\usepackage[utf8]{inputenc} 
\usepackage[T1]{fontenc}    
\usepackage{hyperref}       
\usepackage{url}            
\usepackage{booktabs}       
\usepackage{amsfonts}       
\usepackage{nicefrac}       
\usepackage{microtype}      
\usepackage{xcolor}         
\usepackage{graphicx}

\title{WisdomBot: Tuning Large Language Models with Artificial Intelligence Knowledge}

\author{%
  Jingyuan Chen \\
  College of Education\\
  Zhejiang University\\
  Hangzhou 310058, China \\
  \texttt{jingyuanchen@zju.edu.cn} \\
  \And
  Tao Wu \\
  College of Computer Science and Technology \\
  Zhejiang University\\
  Hangzhou 310058, China \\
  \texttt{twu22@zju.edu.cn} \\
  \AND
  Wei Ji \\
  School of Computing\\
  National University of Singapore  \\
  Singapore 119077, Singapore\\
  \texttt{jiwei@nus.edu.sg} \\
  \And
  Fei Wu \\
  College of Computer Science and Technology \\
  Zhejiang University\\
  Hangzhou 310058, China \\
  \texttt{wufei@zju.edu.cn} \\
}

\begin{document}

\maketitle

\begin{abstract}
Large language models (LLMs) have emerged
as powerful tools in natural language processing (NLP),
showing a promising future of artificial generated
intelligence (AGI). Despite their notable performance in
the general domain, LLMs have remained suboptimal in
the field of education, owing to the unique challenges
presented by this domain, such as the need for more
specialized knowledge, the requirement for personalized
learning experiences, and the necessity for concise
explanations of complex concepts. To address these
issues, this paper presents a novel LLM for education
named WisdomBot, which combines the power of LLMs
with educational theories, enabling their seamless
integration into educational contexts. To be specific, we
harness self-instructed knowledge concepts and
instructions under the guidance of Bloom’s Taxonomy as
training data. To further enhance the accuracy and
professionalism of model’s response on factual questions,
we introduce two key enhancements during inference,
i.e., local knowledge base retrieval augmentation and
search engine retrieval augmentation during inference.
We substantiate the effectiveness of our approach by
applying it to several Chinese LLMs, thereby showcasing
that the fine-tuned models can generate more reliable
and professional responses.
\end{abstract}

\section{Introduction}
In recent years, the field of artificial intelligence (AI)
has witnessed significant advancements, giving rise to a
powerful class of large language models (LLMs). At the
forefront of these language models is GPT-4 \cite{openai2023gpt4}. With an impressive number of parameters in the
trillions, GPT-4 is one of the largest language models
ever created, capable of understanding complex
language patterns and generating responses that are
often indistinguishable from those of a human. At the
same time, open-source communities have developed
several alternatives, including LLaMA \cite{touvron2023llama}. Training on trillions of tokens based on publicly
available data, LLaMA achieves competitive performance with a relatively small number of parameters and
affordable training costs. Thus, it emerges as a valuable
open-source tool for a wide range of applications.

\begin{figure}[t]
  \centering
  \includegraphics[width=1.0\linewidth]{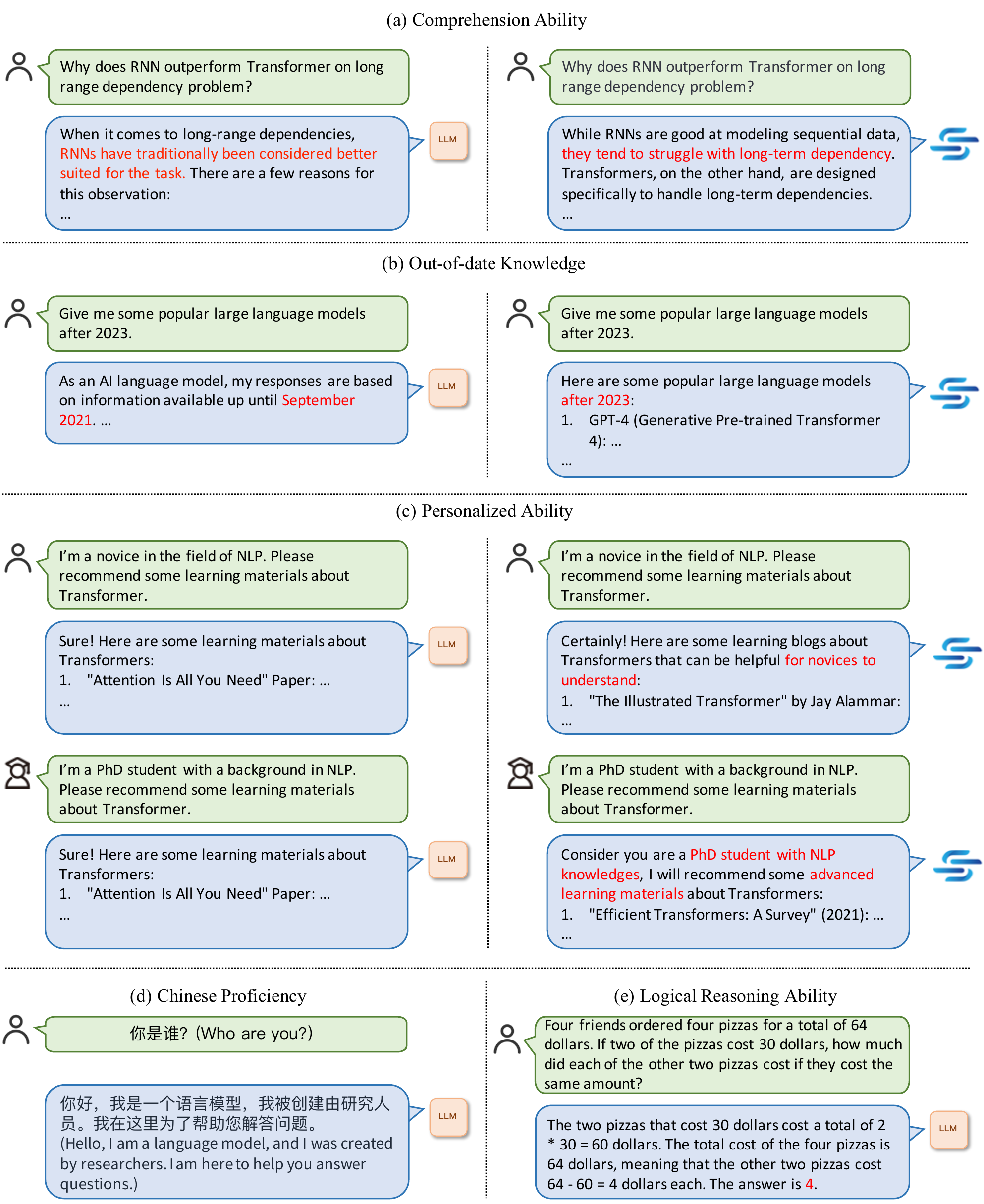}
  \caption{Limitations of general LLMs in education: (a) comprehension ability, (b) out-of-date knowledge, (c) personalized ability, (d) Chinese proficiency, (e) logical reasoning ability.}
  \label{f1}
\end{figure}

While large language models have demonstrated high capacity, they often encounter distinct
challenges when applied to educational tasks. We
outline the key limitations they face as follows and
provide an illustration in Figure \ref{f1}.

First, inadequacy of basic cognitive capacities.
Although LLMs possess exceptional capabilities in
general domain, their fundamental educational
capabilities, such as retention and comprehension,
remain limited due to constraints imposed by the
training data. Primarily, owing to the vastness of
knowledge, LLMs often exhibit constrained
comprehension of specialized expertise that extends
beyond their training data, yielding inaccurate
responses. Additionally, academic knowledge is
incessantly evolving, especially within practical subjects.
The information contained in their training data may
become outdated and obsolete, thus constraining the
basic ability of LLMs to generate factually accurate
responses when faced with inquiries that necessitate up-
to-date awareness of post-training occurrences and
knowledge \cite{cao2021knowledgeable, liu2023evaluating, wang2021can, yang2023chatgpt}.

\begin{figure}[t]
  \centering
  \includegraphics[width=1.0\linewidth]{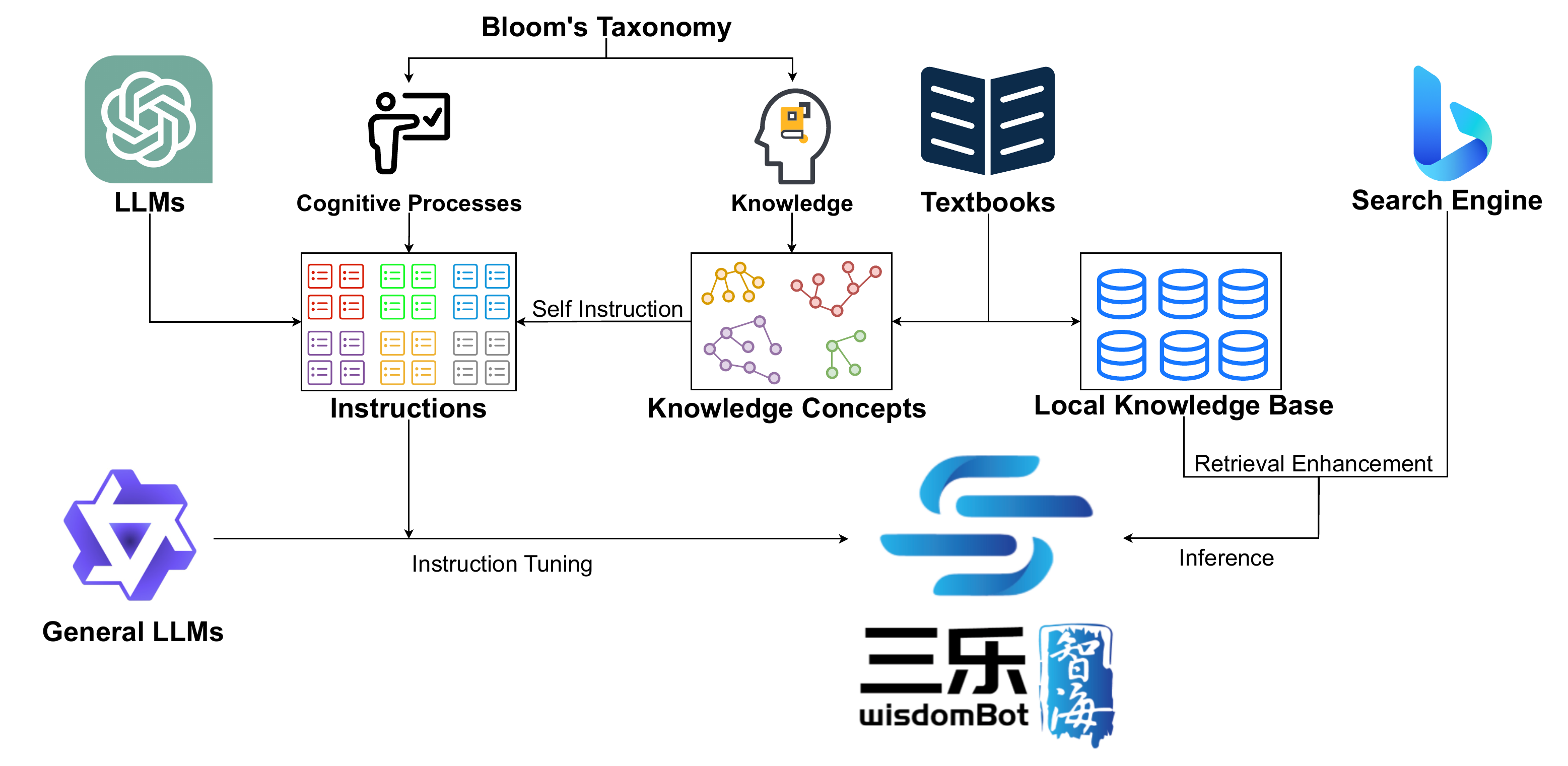}
  \caption{Training pipeline. We collect knowledge concepts and instructions under the guidance of textbooks, Bloom’s Taxonomy and strong LLMs, serving as instruction-tuning data to transform general LLMs to educational LLMs. During inference, we construct a local knowledge base based on the textbook, incorporating search engine capabilities for retrieval enhancement.}
  \label{f2}
\end{figure}

Second, lack of advanced cognitive capacities.
Proficiency in advanced cognitive capacities, such as
analysis, evaluation and innovation, is vital for tackling
challenging tasks. Existing studies have demonstrated
that LLMs lack these capabilities, which can lead to
failures. For instance, Liu et al. \cite{liu2023evaluating} have highlighted
the persisting difficulty in logical reasoning for
ChatGPT and GPT-4, especially when confronted with
unfamiliar data and natural language inference datasets.
Baidoo-Anu and Owusu Ansah et al. \cite{baidoo2023education} discovered that
generative models were limited to generating responses
solely based on patterns present in their training data,
thereby restricting the creativity and originality of their
outputs. Moreover, Baidoo-Anu and Owusu Ansah et al. \cite{baidoo2023education} provided evidence that ChatGPT and other
generative AI models may offer general information
and assistance but lack the ability to personalize
instruction to cater to the specific needs of individual
students. The insufficiency of advanced skills in LLMs
hinders their broader utilization in educational
contexts.

Third, limited Chinese proficiency. While
several large language models, such as LLaMA, have
been made available to the public, their primary focus
has been on English corpora, with limitation in
applicability to other languages. Cui et al. \cite{cui2023chinese}
showed that vocabularies of LLaMA or Alpaca \cite{taori2023stanford} contained only a few hundred Chinese
tokens, substantially hindering their efficiency in
encoding and decoding Chinese text.

The key to rectify the shortcomings and adapt
LLMs to the realm of education lies in the
amalgamation of LLMs with educational theories,
thereby equipping LLMs with varying levels of abilities.
Among a large amount of educational theories, Bloom’
s
Taxonomy, as delineated in \cite{anderson2001revision},
proffers a framework for categorizing the diverse
objectives and proficiencies that educators aspire to
instill into their students. The new taxonomy embraces
a two-dimensional framework encompassing
“knowledge” and “cognitive processes.” “Knowledge”
pertains to the relevant content involved in learning,
while “cognitive processes” refer to the academic
behaviors and manifestations of learning that need to be
mastered. Bloom’s Taxonomy has provided ample
scope for guiding teaching practices and helping
learners progress to higher-order thinking \cite{ramirez2017pedagogy}, henceforth, it ought to be utilized to boost the
abilities of LLMs in the realm of education.

In accordance with Bloom’s Taxonomy, the
authors propose a method that transfers general
domain LLMs into educational field by simultaneously
learning the “knowledge” and “cognitive processes”
dimension in this paper, as illustrated in Figure \ref{f2}.
Firstly, to expand the breadth of the model’s knowledge,
we manually summarize coarse-grained knowledge
concepts drawn from authentic textbooks, and
meticulously utilize strong LLMs to generate fine-
grained knowledge concepts that are aligned with the
“knowledge” category. This approach facilitates a
comprehensive coverage of detailed knowledge
concepts spanning multiple levels and complexities,
which help to enhance the model’s basic cognitive
capacities. Subsequently, we employ self-instruction, as
demonstrated in \cite{wang2023selfinstruct}, to construct over
forty thousands Chinese instructions based on the
“cognitive processes” dimension with educational tasks
including professional knowledge question answering,
test problem generation, and intelligent tutoring. These
instructions not only embody a multitude of
educational capacities at various levels, particularly
those advanced ones, but also encompass all
aforementioned fine-grained knowledge concepts. With
them as instruction-tuning data, the proficiency of
models can be significantly enhanced. Finally, in order
to enhance model’s awareness to the knowledge lies
beyond training data, we utilize two retrieval
augmentation strategies during inference, namely local
knowledge base retrieval and search engine retrieval, to
serve as extra knowledge sources.

We conduct our method on two open-source
Chinese language models, Chinese-LLaMA and Alpaca
\cite{cui2023chinese} and Qwen-7B-Chat \cite{bai2023qwen}.
Experiments have demonstrated the superiority of our
fine-tuned models compared to the original models
across various educational tasks, assessed from a diverse
range of evaluation perspectives. It is worthy of note
that our experiments are conducted specifically within
the domain of Chinese AI instruction. In summary, our
contributions are three-fold:

\begin{itemize}
    \item We devise instructions based on textbooks and the guidance of Bloom’s Taxonomy to transfer general LLMs to educational domain;
    \item We utilize retrieval augmentation strategies during inference to expand the width of the model’s knowledge and enhance the quality of responses to factual inquiries;
    \item We conduct evaluation on various education tasks, demonstrating the superiority of our finetuned models compared to origin models.
\end{itemize}

\section{Related Work}
\subsection{Large Language Models}
LLMs have revolutionized the field of AI and natural
language processing, opening up new possibilities for
human–computer interaction and advancing our
understanding of language and its applications. With
the ability to process and analyse vast amounts of
textual data, large language models have showcased
remarkable capabilities in tasks such as text generation,
question answering, summarization, translation,
sentiment analysis, and more.

Among all the models, ChatGPT \cite{ouyang2022training} and GPT-4 \cite{openai2023gpt4} are two
prominent iterations of large language models
developed by OpenAI. Despite their excellent
performance on general tasks, they are not open-source
models and have a huge amount of parameters, which
hinders personal deployment and research. LLaMA
\cite{touvron2023llama} is an open-source substitute for
GPT, with number of parameters ranging from 7 billion
to 65 billion. Alpaca \cite{taori2023stanford} conducts
instruction tuning on LLaMA with 52K instruction
data, achieving comparable performance with ChatGPT
in English at an affordable cost.

Despite their great performance in English,
LLaMA has shown weakness in Chinese due to the lack
of Chinese corpus in the training data. To tackle this
problem, Chinese-LLaMA-Alpaca \cite{cui2023chinese}
augment LLaMA with capabilities for understanding
and generating Chinese text and its ability to follow
instructions by extending LLaMA’s existing vocabulary
with additional Chinese tokens and further fine-tuning
the model with Chinese instruction datasets. Unlike
Chinese-LLaMA-Alpaca directly fine-tuning on
LLaMA, ChatGLM \cite{zeng2022glm} is a new model
based on the existing GLM structure. With training on
large Chinese–English bilingual corpus and aligning
with human intentions through supervised fine-tuning,
ChatGLM has shown powerful Chinese language
capabilities.

\begin{figure}[t]
  \centering
  \includegraphics[width=1.0\linewidth]{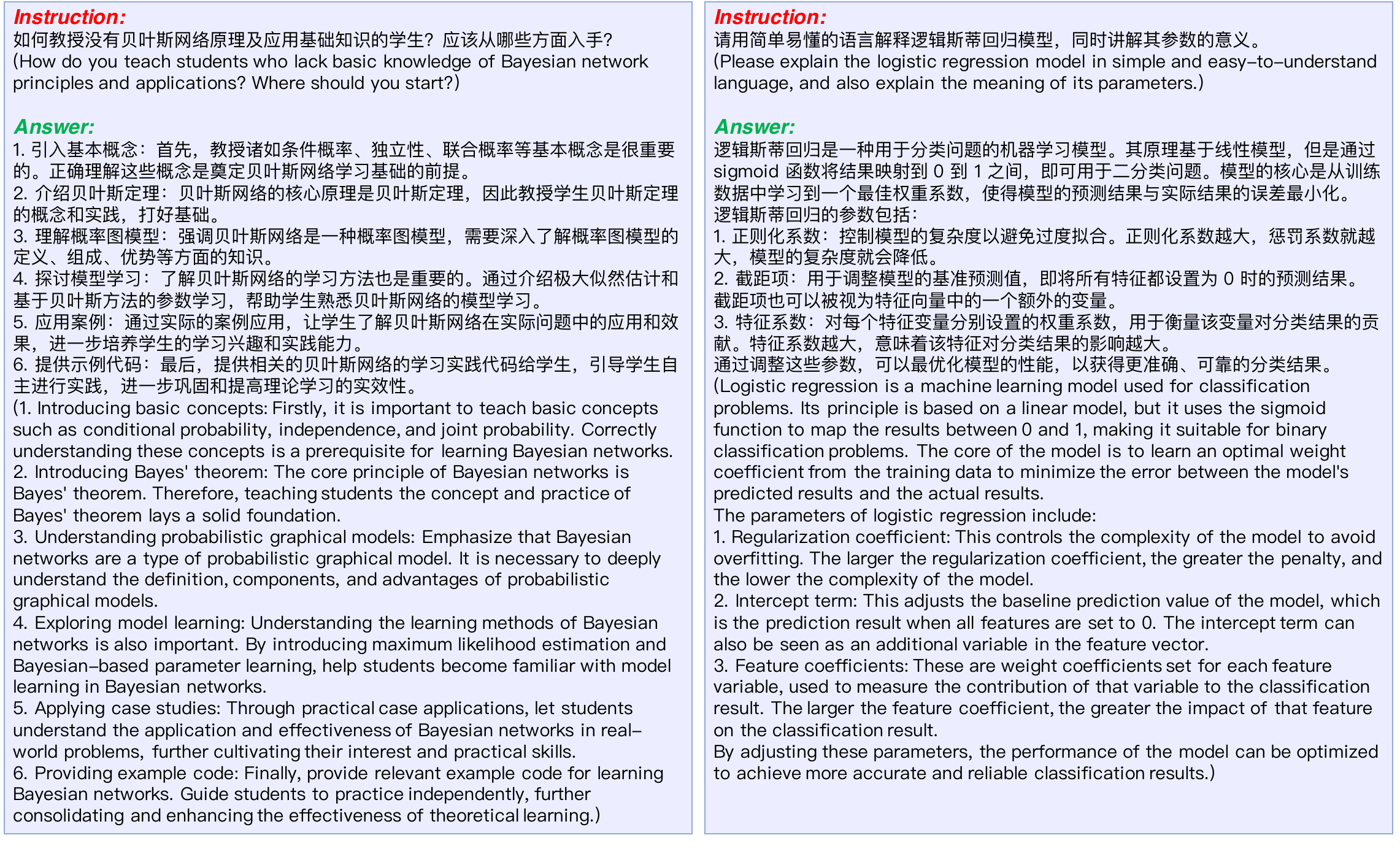}
  \caption{Demonstrations of the dataset.}
  \label{f3}
\end{figure}

\subsection{Bloom’s Taxonomy}
Bloom’s Taxonomy \cite{zeng2022glm} is a widely
recognized framework that categorizes educational
objectives and cognitive processes. It was first
developed in the 1950s by Benjamin Bloom and his
colleagues, and it has since become a fundamental tool
in the field of education. Bloom’s Taxonomy provides a
structured way to understand and organize different
levels of thinking and learning \cite{bloom1956handbook}.

In its 2001 revision, the new taxonomy
embraces a two-dimensional framework encompassing
“knowledge” and “cognitive processes.” Knowledge
pertains to the relevant content involved in learning,
encompassing four categories ranging from concrete to
abstract: factual knowledge, conceptual knowledge,
procedural knowledge, and metacognitive knowledge.
Cognitive processes refer to the academic behaviors and
manifestations of learning that need to be mastered,
including six categories: remember, understand, apply,
analyse, evaluate, and create, which are arranged in
ascending order of cognitive complexity. This theory
has been used to explore the weaknesses of ChatGPT in
the field of education \cite{elsayed2023mitigating}.

\section{Methods}
The details of our finetuning method will be introduced
in this section. Firstly, we collect fine-grained concepts
corresponding to categories in “knowledge” dimension
of Bloom’s Taxonomy. Then the instruction templates
are designed to align with “cognitive processes”
dimension of Bloom’s Taxonomy. By combining
knowledge concepts and instruction templates, we
obtain instruction-output pairs as our training data.
Besides, two retrieval augmentation methods are
developed during inference, enhancing the accuracy
and professionalism of model’
s response.
\subsection{Knowledge Concepts}
Knowledge concepts, as the fundamental unit for
transmitting instructional information in teaching
activities, plays an important role in both teaching and
learning. It can be regarded as the basic component of
subject knowledge and serves as the cornerstone for
constructing a systematic knowledge system. We collect
knowledge concepts with 2 steps, corresponding to
different levels of granularity. Coarse-grained
knowledge concepts have fewer quantities and are easy
to obtain, so we manually extract them from the
textbooks in the first step. In this way, we collect 117
coarse-grained knowledge concepts about AI, which are
used for fine-grained knowledge concepts generation.

In the second step, we aim to collect intricate
fine-grained knowledge concepts that are derived from
the coarse ones. However, acquiring these detailed
concepts manually is a time-consuming task and
requires significant human effort. Therefore, we employ
a self-instruction approach \cite{wang2023selfinstruct} to
acquire them. To be specific, for each coarse concept
and knowledge category within Bloom’s Taxonomy, we
employ ChatGPT to act as an AI learner. We prompt it
to provide a series of questions that it may encounter
during the learning process and to summarize the
corresponding fine-grained knowledge concepts related
to each category. Following careful manual extraction,
cleaning, and filtering of the responses, we obtain a total
of 981 fine-grained concepts and 1,196 questions. These
concepts encompass various levels and diverse subjects.

\subsection{Knowledge-Based Instruction Tuning}
Instruction tuning \cite{wei2021finetuned} is a simple method
to improve the ability of language models to respond to
NLP instructions, demonstrating promising abilities of
language models to perform tasks described purely via
instructions. Inspired by the automatic generation of
instruction, we design templates and construct
instructions using concepts above.

Specifically, we develop 39 distinct templates, in
which fine-grained concepts or questions will be filled.
These templates use three common educational tasks,
i.e., subject knowledge QA, test problem generation,
and intelligent tutoring as carriers, comprehensively
encompassing all the learning abilities described in the
cognitive processes of Bloom’s Taxonomy. Subsequently, these templates are merged with concepts or
questions to formulate original instructions. However,
these instructions often suffer from substandard quality
and a dearth of diversity, which might affect the
performance of the model \cite{wang2023selfinstruct}.
Therefore, we employ ChatGPT to assess the coherence
of each instruction. Only those instructions deemed of
high quality are selected and revised to reduce similarity
with others. Through this process, we acquire a set of
instructions that exhibit exceptional quality. These
instructions are then submitted to human experts
utilizing GPT as an assistant to generate corresponding
answers. Each instruction, along with its respective
answer, is organized following the Stanford Alpaca
\cite{taori2023stanford} style. It consists of indispensable
Instruction and Output fields, while the Input field
remains optional. Ultimately, we obtain 38,784 pairs of
instruction and output, which serve as the foundation
for supervised fine-tuning. We illustrate some data
samples from our dataset in Figure \ref{f3}.

\subsection{Retrieval Enhancement}
As mentioned in \cite{cao2021knowledgeable, liu2023evaluating, wang2021can, yang2023chatgpt}, LLMs have limited
performance in producing factually accurate answers.
To tackle this problem, we utilize two retrieval
enhancement methods during inference, namely local
knowledge base retrieval and search engine retrieval.

Local knowledge base retrieval primarily
addresses factual information contained in textbooks,
which provides advantages when the model responds to
queries involving obscure knowledge or highly
professional language. To establish such a local library,
we follow the standard long chain procedure. Initially,
we import our textbooks to constitute unstructured
textual contents, dividing them into multiple text
chunks using a text splitter. Subsequently, a text
embedding model is employed to transform these text
chunks into a vector space while preserving textual
coherence and similarity. Through this approach, each
query can be converted into the same vector space,
enabling the retrieval of the most similar k text
segments from the textbooks to serve as reference
materials.

To address inquiries that exceed the scope of
textbooks, we employ a search engine to augment our
proficiency. This is accomplished by dispatching each
inquiry to the Azure Bing Search API, which provides
us with a collection of search outcomes serving as
pertinent resources. Whether retrieving information
from a local library or through the search engine, these
resources are amalgamated with the user’
s query,
thereby forming an input for the model to generate a
more accurate, comprehensive, and professional
response. Please note that we set the retrieval
enhancement as an optional feature, which means users
need to decide whether to use the retrieval
enhancement feature according to their needs.

\begin{figure}[t]
  \centering
  \includegraphics[width=1.0\linewidth]{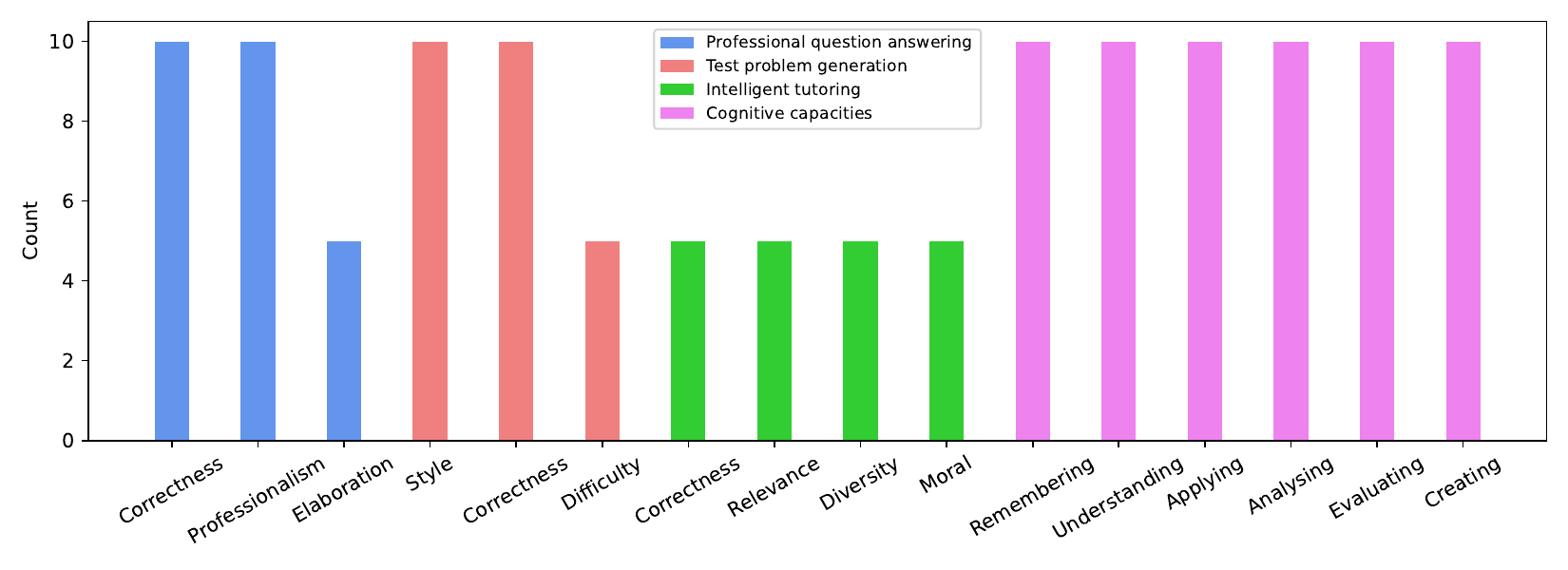}
  \caption{Distribution of test data.}
  \label{f4}
\end{figure}

\section{Experiments}
\subsection{Baselines}
We conduct experiments on two Chinese baselines:
\begin{itemize}
    \item Chinese-LLaMA-Alpaca \cite{cui2023chinese}
continues training on LLaMA with Chinese data. We
conduct our training on Chinese-Alpaca-7B and
Chinese-Alpaca-13B.
\item Qwen \cite{bai2023qwen} is an optimized
dialogue model specifically designed for the Chinese
chatting scenario. We conduct our training on Qwen-
7B-Chat.
\end{itemize}

\subsection{Experiment Detail}
We adopt AdamW optimizer with an initial learning
rate of 2e-5. The models are trained on 8 A100 GPUs
and the batch size of each GPU is set to 16. We use
Low-Rank Adaptation \cite{hu2021lora} training strategy
to reduce training parameters when training on
Chinese-LLaMA-Alpaca. The lora rank of Chinese-
Alpaca-7B is set to 8, while the lora rank of Chinese-
Alpaca-13B is set to 32. For Qwen-7B-Chat, we utilize
full-parameter finetuning. For inference, we set the
temperature to 1, top-p to 0.7, beam size to 1, and
maximum generation length to 1,024.
\subsection{Testset Detail}
\subsubsection{Self-Constructed Dataset}
Our testset consist of 2 parts: educational functions and
cognitive capacities. There are 70 instances in the
educational functions part, encompassing professional
question answering, test problem generation and
intelligent tutoring. Each function is evaluated from
diverse perspectives. The remaining 60 instances are
dedicated to testing cognitive capacities, corresponding
to the 6 cognitive processes described in Bloom’
s Taxonomy. Figure \ref{f4} illustrates the distribution of our
test data.
\subsubsection{Public Dataset: C-Eval}
C-Eval \cite{huang2024c} is a comprehensive Chinese
evaluation suite designed to assess the advanced
knowledge and reasoning abilities of foundation models
in a Chinese context. It includes 13,948 multiple-choice
questions across four difficulty levels (middle school,
high school, college, and vocational) and spans 52
disciplines. C-Eval also features a subset called C-Eval
Hard, focusing on particularly challenging subjects.
Evaluation results show that even the most advanced
models like GPT-4 have significant room for
improvement, highlighting the suite’s ability to
benchmark the capabilities and limitations of current
language models. We conduct evaluation on the
validation set of C-Eval.

\subsection{Results on Self-Constructed Dataset}
We conduct model inference on our self-constructed
dataset, aiming to compare the results generated by the
original LLMs (i.e., Chinese-Alpaca-7B and Chinese-
Alpaca-13B) and our WisdomBot. We utilize both
human and GPT-4 evaluation when comparing
performance, ensuring the accuracy and diversity of
evaluation. For human evaluation, we recruit ten
experts in the field of AI to compare the responses of
two models. For GPT-4 evaluation, we conduct two
evaluation processes for each question. The order of
responses from the two models differs in the prompt of
each evaluation processes because GPT-4 favors
response that come first in the sequence.

\begin{figure}
  \centering
  \includegraphics[width=1.0\linewidth]{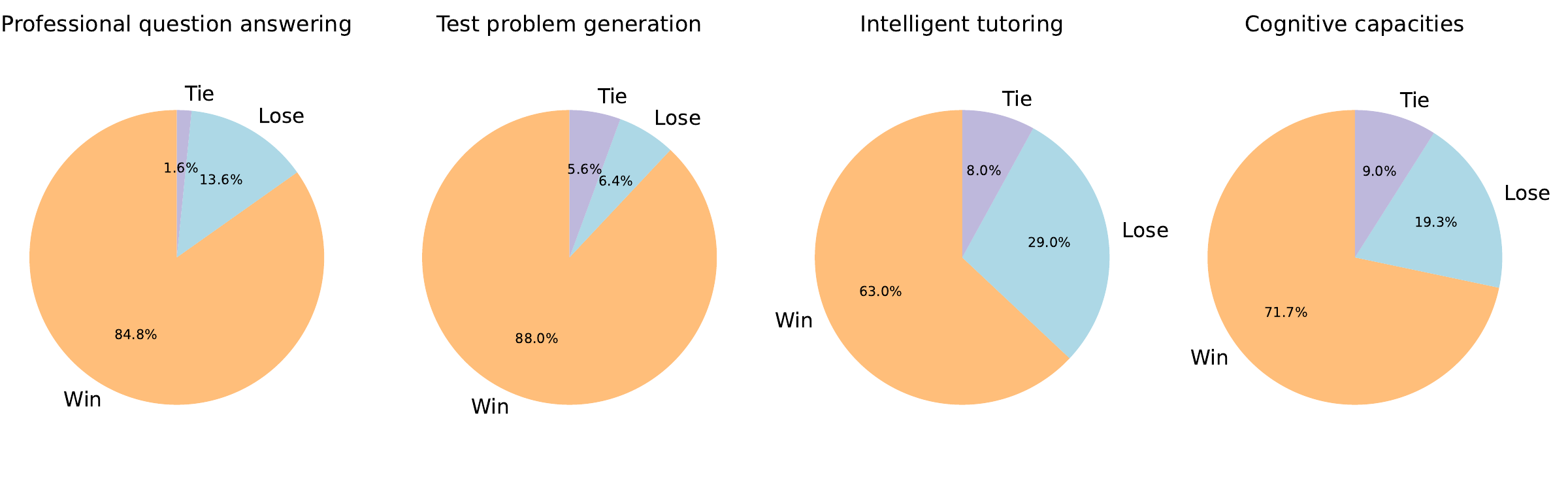}
  \vspace{-2em}
  \caption{Human evaluation of whether WisdomBot outperforms Chinese-Alpaca-7B.}
  \label{f5}
\end{figure}

\begin{figure}
  \centering
  \includegraphics[width=1.0\linewidth]{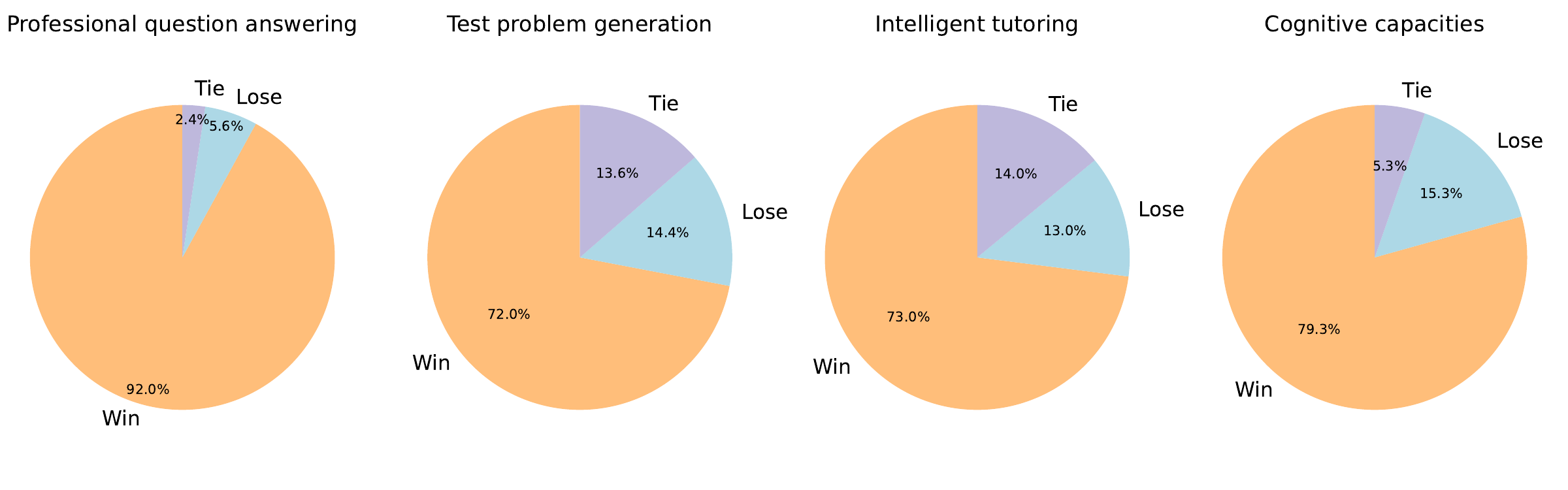}
  \vspace{-2em}
  \caption{Human evaluation of whether WisdomBot outperforms Chinese-Alpaca-13B.}
  \label{f6}
\end{figure}

\begin{figure}
  \centering
  \includegraphics[width=1.0\linewidth]{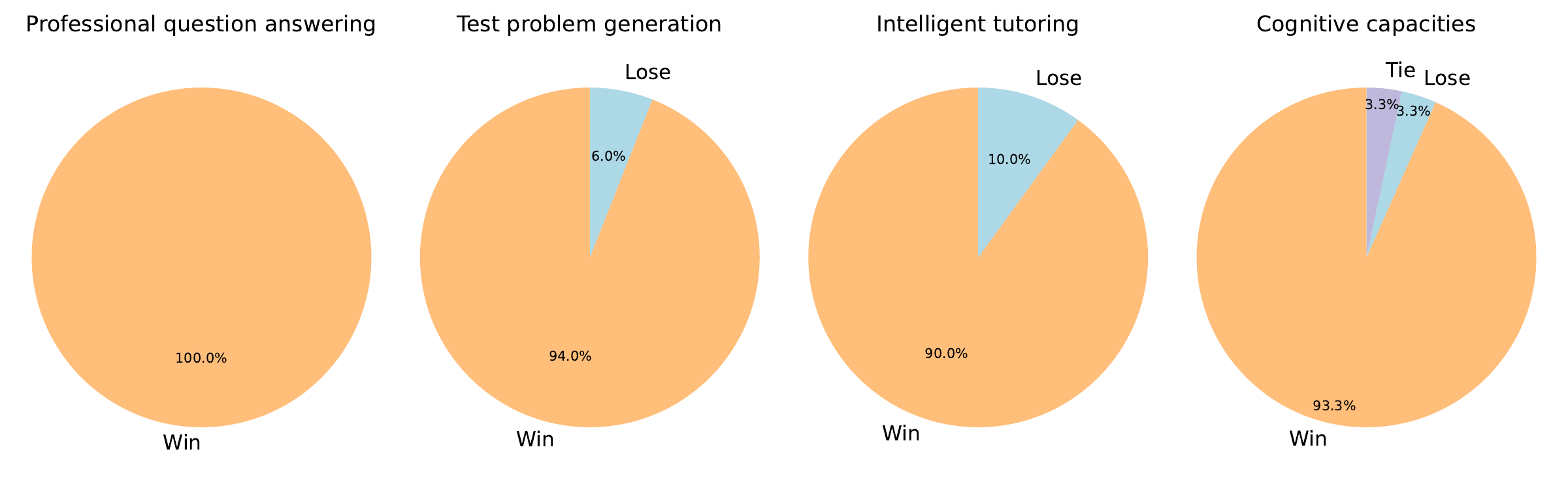}
  \vspace{-2em}
  \caption{GPT-4 evaluation of whether WisdomBot outperforms Chinese-Alpaca-7B.}
  \label{f7}
\end{figure}

\begin{figure}
  \centering
  \includegraphics[width=1.0\linewidth]{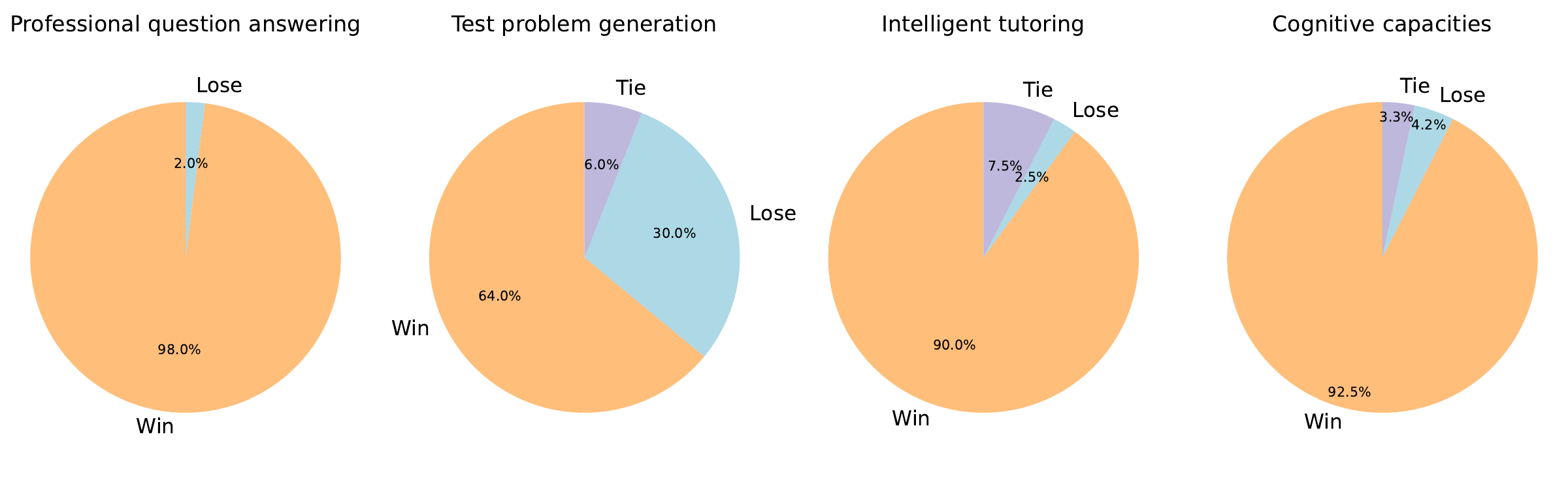}
  \vspace{-2em}
  \caption{GPT-4 evaluation of whether WisdomBot outperforms Chinese-Alpaca-13B.}
  \label{f8}
\end{figure}

The results are shown in Figures \ref{f5}-\ref{f8}. From
these pictures we can observe that for each evaluation
part, WisdomBot has a winning rate of at least 63\%.
WisdomBot even reaches a 100\% winning rate on
professional question answering part comparing with
Chinese-Alpaca-7B. These results demonstrate that
WisdomBot can provide more accurate responses.


\subsection{Results on C-Eval}
We conduct zero-shot evaluation on the validation set
of C-Eval benchmark of all the baselines and
WisdomBot. We list the overall performance of the
three models in Table \ref{t1}, and the performance on each
subset of C-Eval benchmark in Tables \ref{t2}–\ref{t5}. From these
tables we can observe that WisdomBot outperforms the
baselines in most subjects, especially in subjects related
to information and computer science. We attribute the
performance enhancement to our training data, which
is highly relevant to these subjects. For other subjects
such as social science and humanities, WisdomBot does
not exhibit a significant performance decrease. The
performance on “other” subset is even increased
compared with the baselines. The results demonstrate
the superiority of our WisdomBot model.

\begin{table}[h]
  \caption{Results on the validation set of C-Eval benchmark.}
  \label{t1}
  \centering
  \begin{tabular}{ccccccc}
    \toprule
    Model & STEM & Social Science & Humanities & Other & Hard & Average \\
    \midrule
    Chinese-Alpaca-7B & 35.45 & 51.53 & 47.67 & 41.87 & 28.28 & 42.49 \\
    Qwen-7B-Chat & 51.61 & 72.64 &  66.94 & 53.83 & 35.14 & 59.37 \\
    WisdomBot & 59.17 &  72.01 & 65.38 & 54.96 & 49.26 & 62.06 \\
    \bottomrule
  \end{tabular}
\end{table}

\begin{table}[h]
  \caption{Results on the STEM subset within the validation set of C-Eval benchmark.}
  \label{t2}
  \centering\resizebox{1.0\linewidth}{!}{
  \begin{tabular}{ccccc}
    \toprule
    Model & Computer network &Operating system &Computer architecture &College programming \\
    \midrule
    Chinese-Alpaca-7B& 36.84& 52.63& 38.1& 43.24\\
Qwen-7B-Chat& 42.11& 42.11& 52.38& 64.86\\
WisdomBot& 52.63& 57.89& 57.14& 62.16\\
    \bottomrule
    \toprule
    Model &College physics&College chemistry&Advanced mathematics&Probability and statistics\\
    \midrule
    Chinese-Alpaca-7B& 31.58& 16.67& 21.05& 33.33\\
Qwen-7B-Chat& 31.58& 54.17& 10.53& 22.22\\
WisdomBot& 57.89& 58.33& 26.32& 33.33\\
 \bottomrule
\toprule
Model& Discrete mathematics&Electrical engineer&Metrology engineer & High school mathematics\\
\midrule
Chinese-Alpaca-7B& 43.75& 37.84& 50& 16.67\\
Qwen-7B-Chat &18.75 &24.32& 75& 33.33\\
WisdomBot &37.5 &35.14& 70.83& 33.33\\
 \bottomrule
\toprule
Model& High school physics&High school chemistry& High school biology& Middle school mathematics\\
\midrule
Chinese-Alpaca-7B &31.58& 31.58& 42.11& 21.05\\
Qwen-7B-Chat &57.89 &52.63 &73.68 &63.16\\
WisdomBot &78.95& 68.42& 68.42& 63.16\\
 \bottomrule
\toprule
Model &Middle school biology &Middle school physics& Middle school chemistry& Veterinary medicine\\
\midrule
Chinese-Alpaca-7B &47.62 &47.37 &40& 26.09\\
Qwen-7B-Chat &85.71 &84.21 &100 &43.48\\
WisdomBot &90.48 &84.21 &95 &52.17\\
 \bottomrule
  \end{tabular}
  }
\end{table}

\begin{table}[h]
  \caption{Results on the social science subset within the validation set of C-Eval benchmark.}
  \label{t3}
  \centering\resizebox{1.0\linewidth}{!}{
  \begin{tabular}{ccccc}
  \toprule
    Model &College economics& Business administration& Marxism& Mao Zedong Thought\\
\midrule
Chinese-Alpaca-7B& 32.73& 45.45& 52.63& 54.17\\
Qwen-7B-Chat& 45.45& 54.55& 73.68& 75.00\\
WisdomBot& 38.18 &54.55 &84.21& 62.50\\
\bottomrule
\toprule
Model& Education science& Teacher qualification& High school politics& High school geography\\
\midrule
Chinese-Alpaca-7B& 37.93& 59.09& 57.89& 42.11\\
Qwen-7B-Chat& 65.52& 84.09& 94.74& 63.16\\
WisdomBot& 72.4&1 81.82& 94.74 &57.89\\
\bottomrule
\toprule
Model& Middle school politics& Middle school geography\\
\midrule
Chinese-Alpaca-7B& 66.67 &66.67\\
Qwen-7B-Chat& 95.24& 75.00\\
WisdomBot& 90.48 &83.33\\
\bottomrule
  \end{tabular}
  }
\end{table}

\begin{table}[h]
  \caption{Results on the humanities subset within the validation set of C-Eval benchmark.}
  
  \label{t4}
  \centering\resizebox{1.0\linewidth}{!}{
  \begin{tabular}{ccccc}
  \toprule
    Model& Modern Chinese history& Ideological and moral cultivation& Logic& Law\\
    \midrule
Chinese-Alpaca-7B& 52.17& 52.63& 54.55& 20.83\\
Qwen-7B-Chat& 78.26& 84.21& 36.36& 41.67\\
WisdomBot& 69.57& 94.74& 59.09& 37.50\\
\bottomrule
\toprule
Model& Chinese language and literature& Art studies& Professional tour guide& Legal professional\\
\midrule
Chinese-Alpaca-7B& 34.78& 48.48& 51.72& 39.13\\
Qwen-7B-Chat& 56.52& 66.67& 79.31& 43.48\\
WisdomBot& 47.83& 69.70& 68.97& 43.48\\
\bottomrule
\toprule
Model& High school Chinese& High school history& Middle school history\\
\midrule
Chinese-Alpaca-7B& 47.37& 50.00& 72.73\\
Qwen-7B-Chat& 78.95& 80.00& 90.91\\
WisdomBot& 57.89& 75.00& 95.45\\
\bottomrule
  \end{tabular}
  }
\end{table}

\begin{table}[h]
  \caption{Results on the other subset within the validation set of C-Eval benchmark.}
  \label{t5}
  \centering\resizebox{1.0\linewidth}{!}{
  \begin{tabular}{ccccc}
  \toprule
    Model& Civil servant& Sports science& Plant protection& Basic medicine\\
    \midrule
Chinese-Alpaca-7B& 40.43& 57.89& 36.36& 47.37\\
Qwen-7B-Chat& 48.94& 47.37& 68.18& 63.16\\
WisdomBot& 53.19& 52.63& 59.09& 68.42\\
\bottomrule
\toprule
Model& Clinical medicine& Urban and rural planner& Accountant& Fire engineer\\
\midrule
Chinese-Alpaca-7B& 36.36& 52.17& 36.73& 38.71\\
Qwen-7B-Chat& 45.45& 63.04& 51.02& 48.39\\
WisdomBot& 50.00& 60.87& 53.06& 45.16\\
\bottomrule
\toprule
Model& Environmental impact assessment engineer& Tax accountant &Physician\\
\midrule
Chinese-Alpaca-7B& 45.16& 34.69 &34.69\\
Qwen-7B-Chat& 48.39& 53.06& 55.10\\
WisdomBot& 58.06& 44.90& 59.18\\
\bottomrule
  \end{tabular}
  }
\end{table}

\subsection{Advanced Cognitive Ability Comparisons}
We compare WisdomBot with baseline models to
evaluate their advanced cognitive abilities, encompassing creativity, personalized ability, and logical
reasoning ability. We curate 50 test samples for each
ability test. For the creativity and personalization tests,
we ask GPT-4 to score each model’s response on a scale
from 1 to 5, with higher scores indicating stronger
abilities. For the logical reasoning test, we directly assess
the outputs and calculate each model’s accuracy. The
results are reported in Table \ref{t6}, which demonstrates the
superiority of WisdomBot in terms of advanced
cognitive abilities.
\begin{table}[h]
  \caption{Comparisons on retrieval enhancements.}
  \label{t6}
  \centering
  \begin{tabular}{cccc}
            \toprule 
            Model& Creativity&  Personalized ability&  Logical reasoning (\%)\\
            \midrule
Chinese-Alpaca-7B&  2.78&  3.56&  8\\
Qwen-7B-Chat&  2.86&  3.34&  46\\
WisdomBot&  3.28&  3.80 & 52\\
\bottomrule
        \end{tabular}
\end{table}

\subsection{Experiments on Retrieval Enhancement}
We evaluate the effectiveness of two retrieval
enhancements: local knowledge base retrieval and
search engine retrieval. For the knowledge base
retrieval, we curate 50 professional AI questions and
assess the professional level of the answers using GPT-
4. For the search engine retrieval, we curate 30 factual
questions and evaluate the correctness of the answers.
The results, presented in Table \ref{t7}, show that the local
knowledge base retrieval enables the model to generate
more professional answers, while the search engine
retrieval improves the model’s accuracy on factual
questions.

\begin{table}[h]
  \caption{Comparisons on retrieval enhancements.}
  \label{t7}
  \centering
  \begin{tabular}{ccc}
         \toprule 
         Model& Local knowledge base (\%)& Search engine (\%)\\
         \midrule
w/o retrieval& 30& 35\\
w retrieval& 70& 93\\
         \bottomrule
    \end{tabular}
\end{table}


\subsection{Case Study}
\begin{figure}[h]
  \centering
  \includegraphics[width=1.0\linewidth]{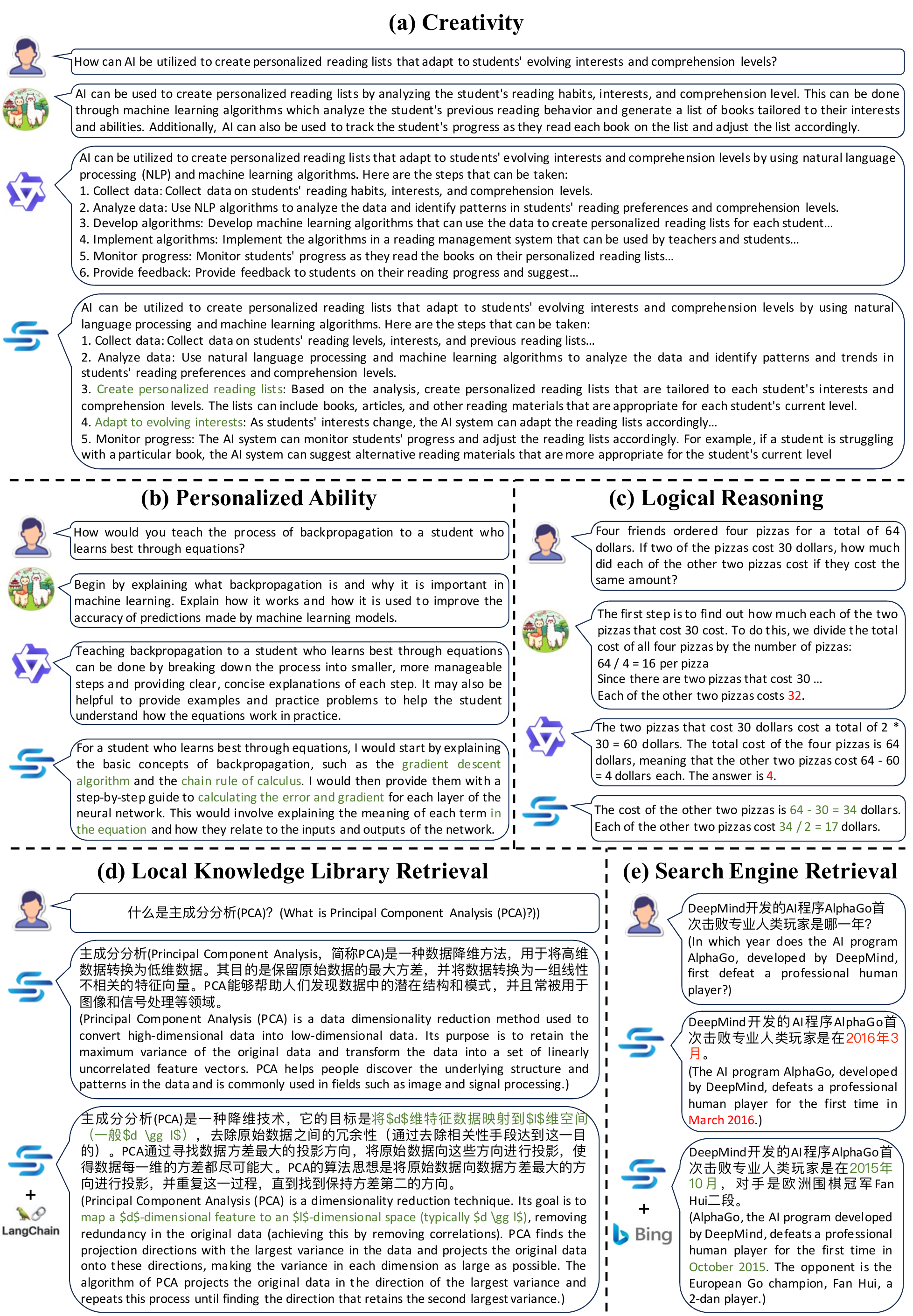}
  \caption{Case examples generated by WisdomBot and baselines: (a) creativity, (b) personalized ability, (c) logical reasoning. WisdomBot with two retrieval enhancement methods: (d) local knowledge library retiveval, (e) search engine retrieval.}
  \label{f9}
\end{figure}

As illustrated in Figure \ref{f9}, WisdomBot demonstrates
enhanced creativity (a), personalized ability (b), and
logical reasoning ability (c) compared to baseline
models. Additionally, local knowledge base retrieval (d)
and search engine retrieval (e) respectively improve the
professional level and accuracy of WisdomBot’
s
response.

\section{Conclusion}
The general large language models lack basic cognitive
abilities and advanced cognitive abilities. We propose a
novel tuning approach, using high-quality textbook-
level corpora as the basis, focusing on knowledge
concepts to construct training data, migrating open-
source large language models to the education field, and
forming the educational large language model
WisdomBot. Experiments show that WisdomBot has
achieved excellent performance in different educational
scenarios and various subjects.

\begin{ack}
This work was supported by the
National Science and Technology Major Project, China
(Grant No. 2022ZD0117104), the National Natural
Science Foundation of China (Grant Nos. 62037001 and
62307032), and the Starry Night Science Fund at
Shanghai Institute for Advanced Study (SN-ZJU-SIAS-
0010).
\end{ack}

\bibliography{main}
\bibliographystyle{unsrt}


\end{document}